\documentclass[10pt,twocolumn,letterpaper]{article}

\usepackage[pagenumbers]{cvpr} 

\definecolor{cvprblue}{rgb}{0.21,0.49,0.74}
\definecolor{bblue}{rgb}{0.0,0.2,0.8}
\usepackage[pagebackref=true,breaklinks=true,colorlinks=true,urlcolor=magenta,bookmarks=false,citecolor=bblue]{hyperref}
\usepackage{multirow}
\usepackage{amssymb}
\usepackage{caption}
\usepackage{xcolor}
\usepackage[dvipsnames]{xcolor}
\usepackage{tikz}
\usepackage{stfloats}

\newcommand{\beginsupplement}{%
        \setcounter{table}{0}
        \renewcommand{\thetable}{S\arabic{table}}%
        \setcounter{figure}{0}
        \renewcommand{\thefigure}{S\arabic{figure}}%
        \setcounter{section}{0}
        \renewcommand{\theequation}{S\arabic{equation}}%
     }

\def\confName{CVPR}
\def\confYear{2026}

\title{
Consistent Instance Field for Dynamic Scene Understanding
}

\author{%
\begin{tabular}{cccc}
Junyi Wu$^{1,2,\dagger}$ & 
Van Nguyen Nguyen$^{2,\ddagger}$ & 
Benjamin Planche$^{2,\ddagger}$ &
Jiachen Tao$^{1,2,\dagger}$ \\
Changchang Sun$^{1}$ &
Zhongpai Gao$^{2}$ &
Zhenghao Zhao$^{1}$ &
Anwesa Choudhuri$^{2}$ \\
\end{tabular}\\
\begin{tabular}{ccc}
Gengyu Zhang$^{1}$ &
Meng Zheng$^{2}$ &
Feiran Wang$^{1}$ \\
Terrence Chen$^{2}$ &
Yan Yan$^{1}$ &
Ziyan Wu$^{2}$
\end{tabular}\\
$^{1}$University of Illinois Chicago, Chicago, IL, USA \quad
$^{2}$United Imaging Intelligence, Boston, MA, USA
}

\newif\ifshowedits
\newcommand{\addeditor}[3]{%
  \definecolor{#1color}{rgb}{#3}
  \expandafter\newcommand\csname #1\endcsname[1]{%
  \ifshowedits
    {\color{#1color} ##1}%
  \else
    {##1}%
  \fi
  }%
  \expandafter\newcommand\csname #1rmk\endcsname[1]{%
  \ifshowedits
    {\color{#1color} {\bf [#2: ##1]}}
  \fi
  }%
  \expandafter\newcommand\csname #1rpl\endcsname[2]{%
  \ifshowedits
    {{\color{#1color} ##1} \sout{##2}}
  \else
    {##1}
  \fi
  }%
}
\begin{document}
\showeditstrue
\newcommand{\ourmethod}{CIF\xspace}

\maketitle

\begingroup
\renewcommand\thefootnote{}\footnotetext{
$^{\dagger}$This work was carried out during the internship of Junyi Wu and Jiachen Tao at 
United Imaging Intelligence, Boston, MA, USA.}
\renewcommand\thefootnote{}\footnotetext{
$^{\ddagger}$Corresponding authors.
}
\endgroup

\begin{abstract}
We introduce Consistent Instance Field, a continuous and probabilistic spatio-temporal representation for dynamic scene understanding.
Unlike prior methods that rely on discrete tracking or view-dependent features, our approach disentangles visibility from persistent object identity by modeling each space–time point with an occupancy probability and a conditional instance distribution. 
To realize this, we introduce a novel instance-embedded representation based on deformable 3D Gaussians, which jointly encode radiance and semantic information and are learned directly from input RGB images and instance masks through differentiable rasterization.
Furthermore, we introduce new mechanisms to calibrate per-Gaussian identities and resample Gaussians toward semantically active regions, ensuring consistent instance representations across space and time. Experiments on HyperNeRF and Neu3D datasets demonstrate that our method significantly outperforms state-of-the-art methods on novel-view panoptic segmentation and open-vocabulary 4D querying tasks.
\end{abstract}

\section{Introduction}
\label{sec:intro}

Dynamic scenes reveal not only how the world changes, but also how its entities persist and interact over time. Understanding them remains a central challenge in computer vision, which goes beyond reconstructing geometry and appearance, and seeks to know what is moving while maintaining temporally consistent semantics.
This capability forms the foundation for a wide range of applications, including augmented/virtual reality \cite{guerroudji20243d, schieber2025semantics, jiang2024vr}, autonomous driving \cite{gu2024conceptgraphs, zhou2024drivinggaussian, xu2025ad}, and robotics \cite{shorinwa2024splat, zhu20243d, li2024object, rashid2023language, su2025medgrpo}.

\begin{figure}[t]
\centering
\newlength{\teaserheight}
\setlength{\teaserheight}{1.4cm}
\setlength{\tabcolsep}{0.5pt}
\setlength{\fboxsep}{0pt}
\setlength{\fboxrule}{2pt} 
\begin{tabular}{@{}cc@{}}
\small{Input RGB} &\hspace{-2.8mm} \;\;\;\;\small{Baseline}\;\;\;\;\;\;\;\;\small{Ours} \\[1pt]
\raisebox{1.3mm}{\raisebox{\dimexpr-\height+\teaserheight\relax}{%
    \begin{tikzpicture}
        \node[anchor=south west, inner sep=0] (image) at (0,0) {%
            \includegraphics[trim=20 0 0 0, clip, height=3.48\teaserheight]{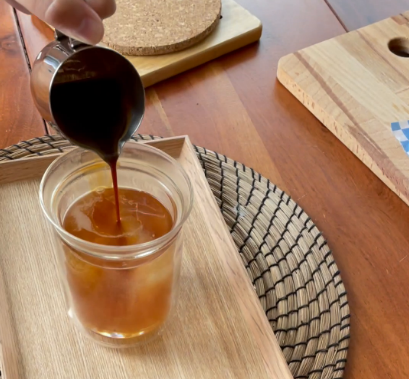}%
        };
        \begin{scope}[x={(image.south east)}, y={(image.north west)}]
            \draw[cyan, line width=1.5pt] (0.01, 0.05) rectangle (0.22, 0.26);
            \draw[orange, line width=1.5pt] (0.82, 0.81) rectangle (0.99, 0.97);
            \draw[ForestGreen, line width=1.5pt] (0.4, 0.8) rectangle (0.6, 0.99);
        \end{scope}
    \end{tikzpicture}%
}} &
\begin{tabular}[t]{@{}cc@{}}
\begin{tikzpicture}
    \node[inner sep=0] (img2) {\fcolorbox{orange}{white}{\includegraphics[viewport=170 145 196 180, clip, width=\teaserheight, height=\teaserheight]{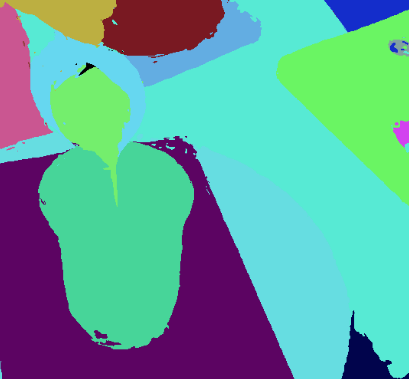}}};
    \node[text=white, font=\fontsize{4.6}{6}\selectfont, align=center, inner sep=1pt, xshift=-2.5mm, fill=black, fill opacity=0.8, text opacity=1, rotate=90] at (img2.west) {\shortstack{weak to inconsistent\\instance supervision}};;
\end{tikzpicture} &
\raisebox{0.7mm}{\fcolorbox{orange}{white}{\includegraphics[viewport=170 145 196 180, clip, width=\teaserheight, height=\teaserheight]{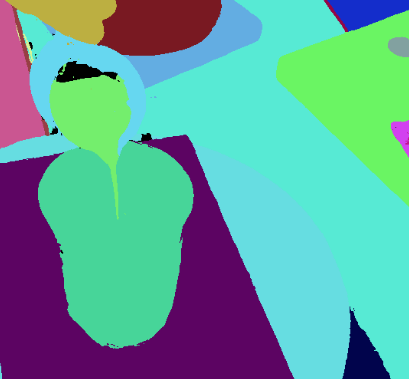}}}
\\
\begin{tikzpicture}
    \node[inner sep=0] (img1) {\fcolorbox{cyan}{white}{\includegraphics[viewport=10 10 60 60, clip, width=\teaserheight, height=\teaserheight]{figures/teaser_baseline.png}}};
    \node[text=white, font=\fontsize{4.6}{6}\selectfont, align=center, inner sep=1pt, xshift=-2.5mm, fill=black, fill opacity=0.8, text opacity=1, rotate=90] at (img2.west) {\shortstack{opacity vs occupancy\\misconception}};
\end{tikzpicture}
 &
 \raisebox{0.7mm}{\fcolorbox{cyan}{white}{\includegraphics[viewport=10 10 60 60, clip, width=\teaserheight, height=\teaserheight]{figures/teaser_ours.png}}}
\\
\begin{tikzpicture}
    \node[inner sep=0] (img1) {\fcolorbox{ForestGreen}{white}{\includegraphics[viewport=85 140 125 180, clip, width=\teaserheight, height=\teaserheight]{figures/teaser_baseline.png}}};
    \node[text=white, font=\fontsize{4.6}{6}\selectfont, align=center, inner sep=1pt, xshift=-2.5mm, fill=black, fill opacity=0.8, text opacity=1, rotate=90] at (img2.west) {\shortstack{semantic sparsity in\\meaningful regions}};
\end{tikzpicture}
 &
 \raisebox{0.7mm}{\fcolorbox{ForestGreen}{white}{\includegraphics[viewport=85 140 125 180, clip, width=\teaserheight, height=\teaserheight]{figures/teaser_ours.png}}}
\\
\end{tabular}
\end{tabular}
\vspace{-5pt}
\vspace{-2mm}
\caption{\textbf{Comparisons with prior work SA4D~\cite{ji2024segment}}. Previous methods like SA4D often rely on view-dependent features with RGB modulation, leading to semantic inconsistencies in dynamic scenes:
\textcolor{orange}{unstable under cross-view instance supervision}, \textcolor{cyan}{confusing color opacity with object occupancy}, and \textcolor{ForestGreen}{underrepresenting semantically meaningful regions}. Our approach formulates a continuous probabilistic field over \emph{existence} and \emph{identity} in space-time, enabling identity modeling beyond visibility cues and adaptive redistribution of Gaussian capacity. This results in a coherent instance field across deformation and changing viewpoints.}
\vspace{-6mm}
\label{fig:teaser}
\end{figure}

Dynamic scene understanding has advanced rapidly with the encouraging progress in 3D representations \cite{mildenhall2021nerf, muller2022instant, kerbl20233d} and large foundation models \cite{radford2021learning, kirillov2023segment}.
Deformable representations, including NeRF-based~\cite{park2021nerfies, pumarola2021d, park2021hypernerf} and Gaussian-based variants~\cite{duisterhof2023md, bae2024per, huang2024sc}, achieve photorealistic reconstructions of motion and appearance, yet they omit explicit modeling of object identity.
Recent works attempted to address this gap by adding vision-language features into 3D representation~\cite{kerr2023lerf, shi2024language, qin2024langsplat, li20254d} or by incorporating 2D mask supervision~\cite{engelmann2024opennerf, wu2024opengaussian, peng2023openscene, li2024sadg, shen2025trace3d}.
However, as shown in Figure~\ref{fig:teaser}, these approaches have inherent limitations: their supervision is mediated through RGB rendering, making them inherently view-dependent. Without explicitly modeling persistent object existence in space-time, they tie identity to radiance and remain vulnerable to visibility bias, which can lead to drifting and underrepresented semantics when objects deform or change appearance across views.

To address this issue, we propose \textbf{Consistent Instance Field} (\textbf{CIF}), a novel framework modeling dynamic scenes using spatio-temporal functions that jointly encode object existence and identity. In our framework, a dynamic scene is represented as a continuous, object-centric field, where each point in space-time is attributed to a persistent entity. This perspective shifts the focus from tracking changing appearances to modeling the persistent composition of objects in 4D: evaluating the field at any 4D location reveals both whether an entity exists and which one it is. To realize this field, we extend deformable Gaussians~\cite{yang2024deformable, luiten2024dynamic}, with each Gaussian acting as a local carrier of geometric, radiometric information, and encodes two probabilistic quantities: an occupancy probability, indicating physical existence, and a conditional identity distribution, specifying the associated instance. These quantities are jointly optimized through differentiable rasterization~\cite{kerbl20233d} of both RGB and semantic fields in our Field-Aware Splatting, enabling coherent photometric and semantic supervision across views and time.

To further align the discrete Gaussian representation with the underlying continuous field, we introduce two key mechanisms. First, \textbf{Instance Identity Estimation} maps input 2D instance masks into our representation via aggregation across views and time. We then incorporate calibration factors to alleviate the visibility bias and refine the identity distributions, converting view-dependent labels into temporally consistent distributions and allowing each Gaussian to maintain stable instance identities. Second, \textbf{Instance-Guided Resampling} adaptively redistributes Gaussian density according to the instance-field signal, concentrating representational capacity around semantically relevant regions while preserving local volumetric balance. Together, these mechanisms enable the representation to self-organize around meaningful entities, achieving coherent geometry, appearance, and identity throughout the 4D scene.

We evaluate our method on standard benchmarks of dynamic scenes, achieving state-of-the-art performance on novel-view panoptic segmentation and open-vocabulary 4D querying tasks.
In summary, our contributions are:
\begin{enumerate}[nosep]
\item We propose Consistent Instance Field (CIF), a continuous and probabilistic spatio-temporal formulation for dynamic scenes that jointly encodes occupancy and instance identity, which is realized in a Gaussian representation and optimized via differentiable rasterization.
\item We introduce two mechanisms, Instance Identity Estimation and Instance-Guided Resampling, that align the discrete Gaussian representation with the continuous field, enabling temporally consistent and semantically coherent instance modeling.
\item We demonstrate that CIF outperforms prior methods, improving average mIoU by +11.4 on HyperNeRF and +5.8 on Neu3D for novel-view panoptic segmentation, while producing sharper boundaries and more accurate instance separation for open-vocabulary 4D querying.
\end{enumerate}

\section{Related Work}
\label{sec:related}

\noindent\textbf{Dynamic Scene Representation.}
Modeling dynamic scenes is a long-standing challenge in 3D vision, often approached by extending static neural representations with deformation fields~\cite{gao2022nerf, chen2024survey}.
Neural Radiance Fields (NeRF)~\cite{mildenhall2021nerf} represent a scene as a continuous volumetric function, later generalized to dynamic settings through deformation fields that map canonical coordinates to their time-dependent counterparts~\cite{park2021nerfies, park2021hypernerf, fang2022fast, jiang2022neuman, song2022pref}.
However, these implicit volumetric models remain computationally heavy and offer limited structural interpretability.
3D Gaussian Splatting~\cite{kerbl20233d} offers an efficient, explicit alternative with real-time rendering~\cite{bao20253d, zhu2025dynamic, wang2025x}.
Its dynamic variants introduce deformation~\cite{wu20244d, yang2024deformable}, trajectories~\cite{li2024spacetime, yoon2025splinegs, wu2025orientation}, or time-conditioned primitives~\cite{duan20244d}, achieving high photometric fidelity but neglecting explicit modeling of instance persistence.
Our work bridges this gap through an instance-consistent formulation that unifies geometry, motion, and identity in a coherent 4D representation.

\begin{figure*}[!t]
  \centering
  \vspace{-2mm}
  \includegraphics[width=1\textwidth]{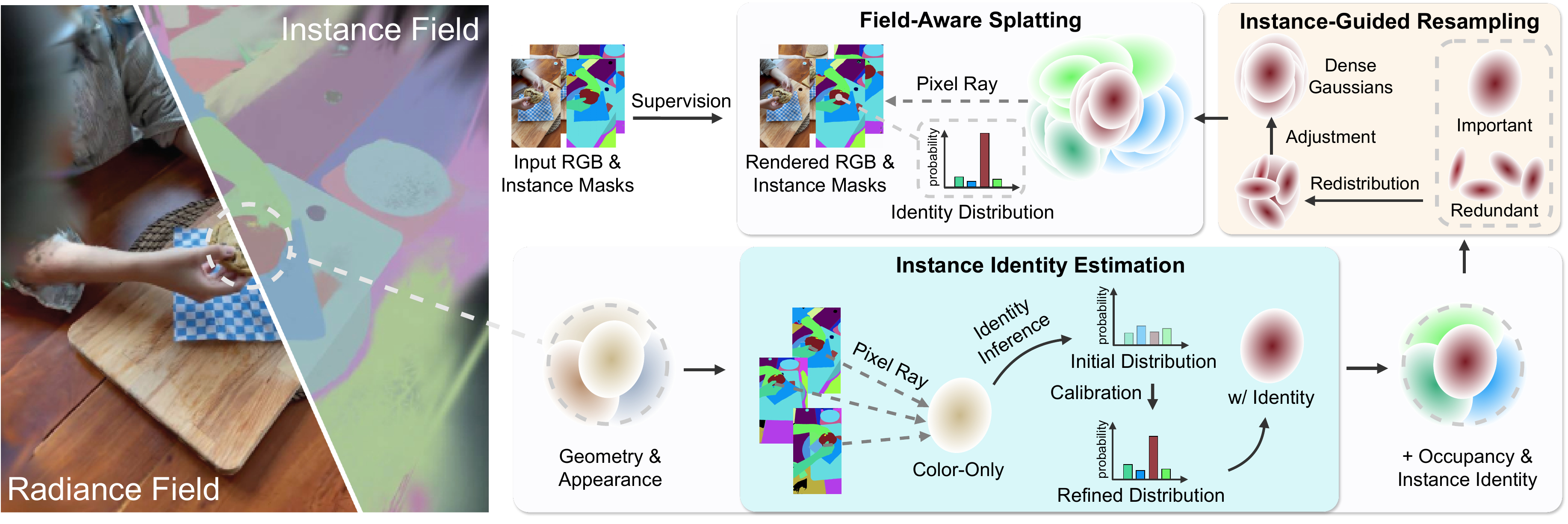}
  \vspace{-7mm}
  \caption{
        \textbf{Overview of our Consistent Instance Field.}
        Our method models each dynamic scene as a continuous 4D Consistent Instance Field that encodes existence and identity distributions (Sec. \ref{sec:method_consistency_formulation}).
        We realize the field as an Instance-Embedded Gaussian Representation, which jointly models geometry, appearance, occupancy, and instance identity (Sec. \ref{sec:method_iegr}).
        \textbf{(Bottom) Instance Identity Estimation.}
        Per-Gaussian identity distributions are inferred by aggregating 2D observations over time and views.
        A learnable calibration then corrects visibility-induced biases (Eqs.~\eqref{vis weight}, \eqref{hat p}, \eqref{final p}), yielding stable identity under occlusion and appearance changes (Sec.~\ref{sec:iie}).
        \textbf{(Right) Instance-Guided Resampling.}
        To align representational capacity with semantic signals, Gaussians with weak instance responses are adaptively relocated toward semantically active regions.
        A volume-conserving adjustment further refines opacity and occupancy (Eqs.~\eqref{vis adj}, \eqref{inst adj}), forming dense object-aligned clusters while preserving radiance and spatial continuity (Sec.~\ref{sec:method:instance_guided_resampling}).
        \textbf{(Top) Field-Aware Splatting.} During rendering, each Gaussian contributes occupancy and identity to per-pixel distributions, supervised with a cross-entropy loss to align Gaussians with the underlying 4D instance field (Sec.~\ref{sec:method:field_aware_splatting}).
  }
  \vspace{-4mm}
  \label{fig:main_figure}
\end{figure*}

\noindent\textbf{Semantic Scene Understanding.}
Beyond reconstruction, recent research has sought to endow 3D representations with semantic understanding.
Point-based approaches \cite{ding2023pla, jatavallabhula2023conceptfusion, liu2023partslip, peng2023openscene, yang2024regionplc} directly embed semantics in point cloud, providing explicit category cues but lacking photometric grounding.
NeRF-based approaches~\cite{engelmann2024opennerf, kerr2023lerf, kobayashi2022decomposing, liu2023weakly} learn semantic and instance fields from vision foundation models~\cite{kirillov2023segment, oquab2023dinov2}, enabling novel-view rendering but remaining implicit and computationally heavy.
More recently, Gaussian-based representations~\cite{kerbl20233d} have emerged as an efficient and interpretable alternative, supporting fine-grained understanding through explicit primitives.
Language-augmented extensions~\cite{jun2025dr, shi2024language, wu2024opengaussian, qin2024langsplat, li20254d, peng20243d} further incorporate vision–language features~\cite{radford2021learning} to enable open-vocabulary querying.
However, most approaches rely on view-dependent semantic cues, resulting in semantics biased toward visible regions and prone to fragmentation under occlusion.
To improve cross-view consistency, recent works~\cite{ji2024segment, cen2025segment, li2024sadg, shen2025trace3d, zhu2025objectgs, ye2024gaussian, lyu2024gaga} leverage 2D masks or video trackers~\cite{cheng2023tracking} to constrain Gaussian semantics, but ambiguous boundaries and occlusions still require heuristic filtering.
In contrast, we introduce a continuous, temporally consistent instance field, where semantics are grounded in the persistence of physical entities, enabling a principled connection with geometry and appearance in dynamic scenes.

\section{Method}\label{sec:method}
In this section, we first introduce our Consistent Instance Field (CIF), including the instance-embedded Gaussian representation, which jointly encodes geometry, radiance, and semantics (Sec.~\ref{sec:method_cif}). We then explain how input 2D masks are propagated through Instance Identity Estimation to establish coherent instance identities (Sec.~\ref{sec:iie}), and how Instance-Guided Resampling adaptively reallocates capacity to preserve the fidelity of the evolving continuous field (Sec.~\ref{sec:method:instance_guided_resampling}). Finally, we describe the joint optimization of all components via differentiable rendering (Sec.~\ref{sec:method:objective}).

\subsection{Consistent Instance Field}\label{sec:method_cif}
\subsubsection{Formulation of Instance Consistency}\label{sec:method_consistency_formulation}
Dynamic scenes can be represented as a continuous 4D field where entities persist, move, and interact through time. To describe this perspective, we introduce the Consistent Instance Field, a spatio–temporal function to capture both their existence and identity over space and time. Formally, let $E\!\in\!\{0,1\}$ be a binary variable indicating whether a 4D location $(\mathbf{x},t)$ is occupied by any physical entity, and $K\!\in\!\mathcal{K}$ a categorical variable representing the instance identity.
The field is expressed as the joint distribution:
\begin{align}
    \gamma(\mathbf{x},t,k) &= P(E{=}1,\,K{=}k \mid \mathbf{x},t) \\
    &= \underbrace{P(E{=}1 \mid \mathbf{x},t)}_{\pi(\mathbf{x},t)} \, \underbrace{P(K{=}k \mid E{=}1, \mathbf{x},t)}_{p(\mathbf{x},t,k)}.
\end{align}
Here, $\pi(\mathbf{x},t){\in}[0,1]$ is the probability that the location $(\mathbf{x},t)$ is occupied by objects, while $p(\mathbf{x},t,k)$ is a conditional distribution over instance identities with $\sum_k p(\mathbf{x},t,k){=}1$.
Each value $\gamma(\mathbf{x},t,k)$ therefore quantifies the probability that instance $k$ occupies the space-time coordinate $(\mathbf{x},t)$.

This decomposition separates the \emph{existence of matter} from the \emph{persistence of identity}: the occupancy $\pi$ models spatial-temporal continuity of physical presence, while the conditional identity $p$ maintains consistent affiliation across deformation and motion. This allows our formulation to capture not only where geometry resides but also which entity it belongs to. The Consistent Instance Field thus defines an object-centric partition of space-time, where low-entropy regions signify stable ownership and higher-entropy zones near interactions represent softly shared boundaries.

\subsubsection{Instance-Embedded Gaussian Representation}\label{sec:method_iegr}
To ground the Consistent Instance Field into an explicit framework, we adopt an Instance-Embedded Gaussian Representation, which approximates the continuous field $\gamma(\mathbf{x},t,k)$ through a finite set of spatially localized primitives.
Building upon Gaussian-based dynamic scene representations \cite{luiten2024dynamic, wu20244d}, each primitive is extended to not only encode geometry and radiance but also instance semantics.
Through this integration, the scene can be represented as a set of Gaussians
\[
\mathcal{G}{=}\{ g_i{=}(\mathbf{x}_i, \mathbf{R}_i, \mathbf{s}_i, \mathbf{c}_i, \alpha_i, \pi_i, p_i^1,\dots,p_i^K) \}_{i=1}^N,
\]
where $\mathbf{x}_i{\in}\mathbb{R}^3$ denotes the center, $\mathbf{R}_i{\in}SO(3)$ the rotation, $\mathbf{s}_i{\in}\mathbb{R}_+^3$ the scale, $\mathbf{c}_i{\in}\mathbb{R}^3$ the color, $\alpha_i{\in}[0,1]$ the opacity, and $(\pi_i, p_i^1,\dots,p_i^K)$ the probability of existence and identity distribution over $K$ instances defined in Sec.~\ref{sec:method_consistency_formulation}.
Representation of dynamic scenes is modeled by an additional time-conditioned MLP \cite{yang2024deformable, wu20244d}, which produces smooth trajectories for each Gaussian and modulates its parameters over time. Each primitive carries instance coefficients:
\begin{equation}
    \pi_i \approx \pi(\mathbf{x}_i(t),t),
    \quad
    p_i^k \approx p(\mathbf{x}_i(t),t,k).
\end{equation}
These parameters connect the continuous field to discrete primitives to track both object existence and identity. As each Gaussian evolves under the deformation field, it preserves its semantic identity, propagating instance information in the dynamic scene while maintaining local consistency across objects.

\subsubsection{Field-Aware Splatting}\label{sec:method:field_aware_splatting}
With the Instance-Embedded Gaussian Representation established, we detail how the Consistent Instance Field guides the rendering process.
In our formulation, splatting~\cite{kerbl20233d, peng20243d} can marginalize the 4D occupancy-identity distribution $\gamma(\mathbf{x},t,k)$ along camera rays, converting the geometric and semantic attributes of Gaussians into view-dependent color and instance maps.
For appearance, we adopt the standard alpha-compositing rule~\cite{kerbl20233d}. 
The color at pixel $(u,v)$ and time $t$ is given by:
$
    \mathbf{C}(u,v,t)
    {=} \sum_i T_i(u,v,t)\,\alpha_i(t)\,P_i(u,v,t)\,\mathbf{c}_i(t),
$
where $P_i(u,v,t)$ is the Gaussian weight obtained by projecting the mean and covariance of $g_i$, and
$
    T_i(u,v,t)
    {=} \prod_{j<i} \bigl(1 - \alpha_j(t)\,P_j(u,v,t)\bigr)
$
is the accumulated transmittance.

To incorporate instance information, we extend this process to render the marginal instance identity map:
\begin{equation}
    \mathbf{M}_k(u,v,t) = \sum_i T_i^{\text{inst}}(u,v,t)\,\pi_i\,P_i(u,v,t)\,p_i^k,
\label{inst render}
\end{equation}
with instance transmittance
$
T_i^{\text{inst}}(u,v,t) = \prod_{j<i} \bigl(1 - \pi_j\,P_j(u,v,t)\bigr).
$
Intuitively, $\pi_i$ governs the spatial support of each Gaussian in the 4D field, while $p_i^k$ encodes its semantic affiliation to instance $k$.
The rendered map $\mathbf{M}_k$ therefore represents a soft assignment of each pixel to instance $k$, jointly shaped by geometry, occupancy, and identity.

\subsection{Instance Identity Estimation}\label{sec:iie}
The Consistent Instance Field defines a conditional distribution over instance identities, which can be empirically grounded in real observations and adapt to scene dynamics.
We achieve this through Instance Identity Estimation, which links 2D input instance masks to the Gaussian representation through probabilistic inference.
This step converts pixel-level supervision into consistent identity associations, as illustrated in Figure \ref{fig:main_figure} (Bottom).

\vspace{1mm}
\noindent\textbf{Inferring Gaussian Identities from 2D Masks}.
We first obtain per-frame instance masks using DEVA~\cite{cheng2023tracking}, which provides temporally consistent instance segmentation. 
To relate these masks to the Gaussian primitives, we exploit the splatting process to infer the fractional contribution of each Gaussian to image formation~\cite{jun2025dr, xiong2025splat}.
For each pixel $(u,v,t)$, we define the normalized rendering weight:
\begin{equation}
w_i(u,v,t) =
\frac{T_i(u,v,t)\,\alpha_i(t)\,P_i(u,v,t)}
{\sum_j T_j(u,v,t)\,\alpha_j(t)\,P_j(u,v,t)},
\end{equation}
representing the fraction of the pixel's color that is explained by $g_i$ from a posterior perspective.
Aggregating these weights over all pixels and time frames yields an empirical estimate of how often each Gaussian participates in explaining instance $k$:
\begin{equation}
    \tilde{p}_i^k = \frac{\sum_{t,(u,v)} \mathbf{1}\!\left[\mathbf{M}_t(u,v)=k\right]\; w_i(u,v,t)}{\sum_{t,(u,v)} w_i(u,v,t)},
\label{vis weight}
\end{equation}
\begin{equation}
\hat{p}_i^k = \frac{\tilde{p}_i^k}{\sum_{k'} \tilde{p}_i^{k'}}.
\label{hat p}
\end{equation}
The resulting $\hat{p}_i^k$ serves as an initialization of the Gaussian's identity distribution, providing a visibility-weighted estimate of instance affiliation.

\vspace{1mm}
\noindent\textbf{Visibility Bias Calibration.}
Since the rendered weights depend on photometric transmittance, frequently visible or well-illuminated regions may dominate supervision, while occluded or low-contrast regions are underrepresented.
To compensate for this imbalance, we introduce learnable calibration factors $m_i^k > 0$ that rescale the initial distribution:
\begin{equation}
p_i^k
=
\frac{\hat{p}_i^k\,m_i^k}{\sum_{k'} \hat{p}_i^{k'}\,m_i^{k'}}.
\label{final p}
\end{equation}
These factors can absorb residual discrepancies between the visibility-biased estimate and the underlying 4D instance field. 
They are optimized jointly with all other Gaussian parameters through the instance-field rendering in Eq.~\eqref{inst render}.
During training, gradients propagate through both occupancy $\pi_i$ and calibrated identity $p_i^k$, enabling the model to refine instance assignments from purely appearance-driven initialization toward temporally consistent, geometry-aware identity estimation across the dynamic scene.

\subsection{Instance-Guided Resampling}\label{sec:method:instance_guided_resampling}
While the Consistent Instance Field provides a continuous description of spatial occupancy and instance identity, its discretization through a finite set of Gaussians can lead to suboptimal capacity allocation. 
Regions carrying strong semantic signals may be underrepresented, whereas uninformative or background areas may retain redundant primitives~\cite{kheradmand20243d}, deviating from the underlying field.
To address this imbalance, we introduce Instance-Guided Resampling, an adaptive refinement mechanism that reallocates Gaussians according to the strength of their instance affiliation.
We illustrate this process in Figure~\ref{fig:main_figure} (right).

\vspace{1mm}
\noindent\textbf{Adaptive Redistribution.}
For a given instance $k$, we define the instance response of each Gaussian as 
$\gamma_i^k = \pi_i p_i^k$, 
which jointly measures its space-time occupancy $\pi_i$ and semantic affinity $p_i^k$. 
We leverage this signal to construct two complementary sampling distributions:
\begin{equation}
P_{\text{weak}}(i|k) \propto (\gamma_i^k)^{-1},
\qquad
P_{\text{strong}}(i|k) \propto \gamma_i^k,
\end{equation}
where $\gamma_i^k$ is clamped to $[\epsilon,1]$ for numerical stability before being normalized into probability distributions.

Intuitively, $P_{\text{strong}}$ favors Gaussians that strongly support instance $k$, while $P_{\text{weak}}$ emphasizes those contributing weakly or redundantly.
We then sample a weak-strong pair $(w,s)$ from $P_{\text{weak}}$ and $P_{\text{strong}}$ within the same instance, and treat $g_s$ as a \emph{source} primitive from which a new replica is spawned by reinitializing $g_w$ near $g_s$ and inheriting its geometric and semantic attributes.
This operation transfers representational capacity to semantically active regions, encouraging the discrete ensemble to align more closely with the underlying 4D field distribution.

\vspace{1mm}
\noindent\textbf{Volume-Conserving Adjustment.}
Naively replicating Gaussians without regulation may lead to local over-saturation, where multiple primitives overlap and artificially inflate both photometric contribution and semantic confidence.
Let $g_{\text{src}}$ be the Gaussian selected for replication and $n$ the number of replicas it has already produced.
To prevent such optimization instabilities~\cite{liu2025deformable}, we apply a volume-conserving adjustment to both opacity and occupancy for the source Gaussian and all of its new replicas:
\begin{equation}
    \alpha_{\text{src}}^{\text{new}} = \alpha^{\text{new}} = 1 - (1 - \alpha_{\text{src}})^{1/(n+1)},
\label{vis adj}
\end{equation}
\begin{equation}
    \pi_{\text{src}}^{\text{new}} = \pi^{\text{new}} = 1 - (1 - \pi_{\text{src}})^{1/(n+1)}.
\label{inst adj}
\end{equation}
This adjustment locally preserves the effective volumetric contribution of the Gaussian cluster after redistribution, preventing semantic drift or radiance inflation, and enabling the representation to refine adaptively while maintaining visual fidelity and semantic coherence.

\subsection{Training Objective}
\label{sec:method:objective}
All components of our framework are jointly optimized through differentiable field rendering.
At each training iteration, both RGB images and instance maps are rendered as described in Sec.~\ref{sec:method:field_aware_splatting}. The overall training objective combines photometric loss $\mathcal{L}_{\text{rgb}}$ and semantic loss $\mathcal{L}_{\text{inst}}$, encouraging the model to produce temporally consistent, geometry-aware instance assignments. Formally, the loss is defined as:
\begin{equation}
\mathcal{L} = \mathcal{L}_{\text{rgb}} + \lambda_{\text{inst}}\,\mathcal{L}_{\text{inst}},
\end{equation}
where $\lambda_{\text{inst}}$ balances the contribution of two different losses. For photometric loss $\mathcal{L}_{\text{rgb}}$, we use $\ell_1$ between the rendered and ground-truth RGB images:
$
\mathcal{L}_{\text{rgb}} {=} \|\mathbf{C}^{\text{rendered}} {-} \mathbf{C}^{\text{gt}}\|_1.
$ For semantic supervision, the instance loss $\mathcal{L}_{\text{inst}}$ is defined as the cross-entropy between the rendered and ground-truth instance masks:
$
\mathcal{L}_{\text{inst}}
= - \sum_{u,v,t}\sum_{k} 
\mathbf{M}_{k}^{\text{gt}}(u,v,t)\,
\log \mathbf{M}_{k}^{\text{rendered}}(u,v,t).
$

\section{Experiments}
\label{sec:exp}
In this section, we first describe the experimental setup for two dynamic scene understanding tasks: novel-view panoptic segmentation, which jointly evaluates spatial accuracy and temporal consistency of instance identities, and open-vocabulary 4D querying, which evaluates the ability to retrieve instances in space-time based on textual descriptions (Sec.~\ref{sec:exp_setup}). We then compare our proposed method, \ourmethod, with the state-of-the-art~\cite{peng20243d, ji2024segment, jun2025dr, shen2025trace3d} on standard benchmarks, HyperNeRF~\cite{park2021hypernerf} and Neu3D~\cite{li2022neural} (Sec.~\ref{sec:exp_compare_sota}). Finally, we present ablation studies to evaluate the impact of different design choices in our method (Sec.~\ref{sec:exp_ablation}).

\subsection{Experimental Setup}
\noindent\textbf{Evaluation Datasets.}
\label{sec:exp_setup}
We evaluate our proposed method on both monocular and multi-view dynamic scene datasets: HyperNeRF~\cite{park2021hypernerf}, which provides monocular videos of complex human-object interactions captured by a moving camera; and Neu3D~\cite{li2022neural}, which contains synchronized multi-view recordings of complex scenes. As in previous work~\cite{ji2024segment, li20254d}, we use DEVA~\cite{cheng2023tracking} to obtain the ground-truth instance masks as these are not provided in either the original dataset. In the multi-view dataset, to synchronize the instance identities across view, we treat the spatial multi-view dimension as inter-temporal and merge all views into a single pseudo-monocular sequence.
Nonetheless, segmentation can still be inconsistent due to occlusions, \ie, some objects may be fully occluded in certain views, which is inherently ill-posed.
Therefore, to avoid cross-view inconsistencies in the ground truth, we only consider instances that remain visible across all camera views. More details are provided in the supplementary material.

\noindent\textbf{Evaluation Metrics.}
To evaluate the novel-view panoptic segmentation task, we follow previous works~\cite{li2024sadg, ji2024segment, jun2025dr, ye2024gaussian, li20254d}, rendering the Gaussians from novel views and computing three standard metrics to assess the quality of the rendered instance masks, including:
\textbf{(i) mAcc-pix}: the mean pixel accuracy within instance masks, computed as the average ratio of correctly labeled pixels to total pixels across all frames;
\textbf{(ii) mAcc-inst}: the mean instance accuracy, obtained by averaging per-instance pixel-wise prediction accuracies across views, thus treating each instance equally regardless of size; \textbf{(iii) mIoU}: the mean Intersection-over-Union between predicted and reference instance masks, averaged over all instances and frames.
For the open-vocabulary 4D querying task, we follow prior work \cite{ye2024gaussian, li2024sadg} and, given a text prompt, use Grounded DINO~\cite{liu2024grounding} to generate 2D masks that are reprojected into 3D to obtain the corresponding Gaussians, which we render in the test views.

\noindent\textbf{Baselines.}
We evaluate against recent state-of-the-art methods for scene understanding.
Dr. Splat \cite{jun2025dr} and VLGS \cite{peng20243d} incorporate semantics into 3DGS, while Trace3D \cite{shen2025trace3d} and SA4D \cite{ji2024segment} perform instance segmentation in 3D and 4D domains, respectively.
We also include 4D LangSplat \cite{li20254d}, which is the recent state-of-the-art for open-vocabulary querying in dynamic scenes.
All results on HyperNeRF and Neu3D are reproduced under our unified evaluation protocol, as the original implementations are either unavailable or limited to static 3D scenes. We adapt Dr. Splat and Trace3D for 4D data and re-implement SA4D and VLGS based on their papers.

\begin{table*}[t]
    \setlength{\tabcolsep}{3pt} 
    \centering
    \small
    \caption{
        \textbf{Quantitative comparison of our method with the state-of-the-art on novel-view panoptic segmentation using the HyperNeRF~\cite{park2021hypernerf} dataset.}
        We report mAcc-pix, mAcc-inst, and mIoU metrics.
        The \colorbox{red!25}{best}, \colorbox{orange!25}{second best}, and \colorbox{yellow!25}{third best} results are highlighted.
    }
    \vspace{-2mm}
    \label{Table: hypernerf}
    \resizebox{0.9\textwidth}{!}{%
    \small
    \begin{tabular}{l ccc c ccc c ccc}
        \toprule
        & \multicolumn{3}{c}{americano}
        & & \multicolumn{3}{c}{split-cookie}
        & & \multicolumn{3}{c}{chickchicken} \\
        \cmidrule{2-4} \cmidrule{6-8} \cmidrule{10-12}
        Method
        & mAcc-pix & mAcc-inst & mIoU
        & & mAcc-pix & mAcc-inst & mIoU
        & & mAcc-pix & mAcc-inst & mIoU \\
        \midrule
        Dr. Splat~\cite{jun2025dr} & 87.22 & 62.84 & 56.13 && 89.44 & 63.10 & 56.85 && 84.96 & 57.45 & 52.86 \\
        Trace3D~\cite{shen2025trace3d} & 95.81 & 77.96 & 72.33 && 93.79 & 75.87 & 68.80 && 92.29 & 63.88 & 56.68 \\
        VLGS~\cite{peng20243d} & \colorbox{orange!25}{97.56} & \colorbox{orange!25}{86.38} & \colorbox{orange!25}{82.50} && \colorbox{orange!25}{96.35} & \colorbox{orange!25}{80.67} & \colorbox{orange!25}{76.08} && \colorbox{orange!25}{95.38} & \colorbox{orange!25}{68.68} & \colorbox{orange!25}{63.39} \\
        SA4D~\cite{ji2024segment} & \colorbox{yellow!25}{96.45} & \colorbox{yellow!25}{79.27} & \colorbox{yellow!25}{74.54} && \colorbox{yellow!25}{95.50} & \colorbox{yellow!25}{76.95} & \colorbox{yellow!25}{72.60} && \colorbox{yellow!25}{94.94} & \colorbox{yellow!25}{68.55} & \colorbox{yellow!25}{62.52} \\
        \textbf{Ours} & \colorbox{red!25}{98.40} & \colorbox{red!25}{91.73} & \colorbox{red!25}{87.48} && \colorbox{red!25}{97.93} & \colorbox{red!25}{90.40} & \colorbox{red!25}{86.03} && \colorbox{red!25}{96.50} & \colorbox{red!25}{82.31} & \colorbox{red!25}{75.07} \\
        \midrule
        & \multicolumn{3}{c}{espresso}
        & & \multicolumn{3}{c}{keyboard}
        & & \multicolumn{3}{c}{torchocolate} \\
        \cmidrule{2-4} \cmidrule{6-8} \cmidrule{10-12}
        Method
        & mAcc-pix & mAcc-inst & mIoU
        & & mAcc-pix & mAcc-inst & mIoU
        & & mAcc-pix & mAcc-inst & mIoU \\
        \midrule
        Dr. Splat~\cite{jun2025dr} & 87.22 & 62.84 & 56.13 && 89.44 & 63.10 & 56.85 && 84.96 & 57.45 & 52.86 \\
        Trace3D~\cite{shen2025trace3d} & 85.11 & 60.03 & 51.59 && 86.40 & 66.00 & 55.43 && 87.54 & 64.15 & 57.39 \\
        VLGS~\cite{peng20243d} & \colorbox{yellow!25}{91.71} & \colorbox{yellow!25}{70.25} & \colorbox{yellow!25}{62.16} && \colorbox{orange!25}{93.79} & \colorbox{yellow!25}{72.80} & \colorbox{yellow!25}{64.51} && \colorbox{yellow!25}{91.09} & \colorbox{yellow!25}{64.70} & \colorbox{yellow!25}{59.66} \\
        SA4D~\cite{ji2024segment} & \colorbox{orange!25}{91.97} & \colorbox{orange!25}{73.00} & \colorbox{orange!25}{64.76} && \colorbox{yellow!25}{93.70} & \colorbox{orange!25}{73.29} & \colorbox{orange!25}{64.94} && \colorbox{orange!25}{92.87} & \colorbox{orange!25}{72.10} & \colorbox{orange!25}{65.44} \\
        \textbf{Ours} & \colorbox{red!25}{94.73} & \colorbox{red!25}{84.03} & \colorbox{red!25}{75.80} && \colorbox{red!25}{94.73} & \colorbox{red!25}{80.17} & \colorbox{red!25}{71.87} && \colorbox{red!25}{96.10} & \colorbox{red!25}{85.52} & \colorbox{red!25}{80.55} \\
        \bottomrule
    \end{tabular}
    }
\end{table*}

\begin{figure*}[!t]
  \centering
  \vspace{-1mm}
  \includegraphics[width=1\textwidth]{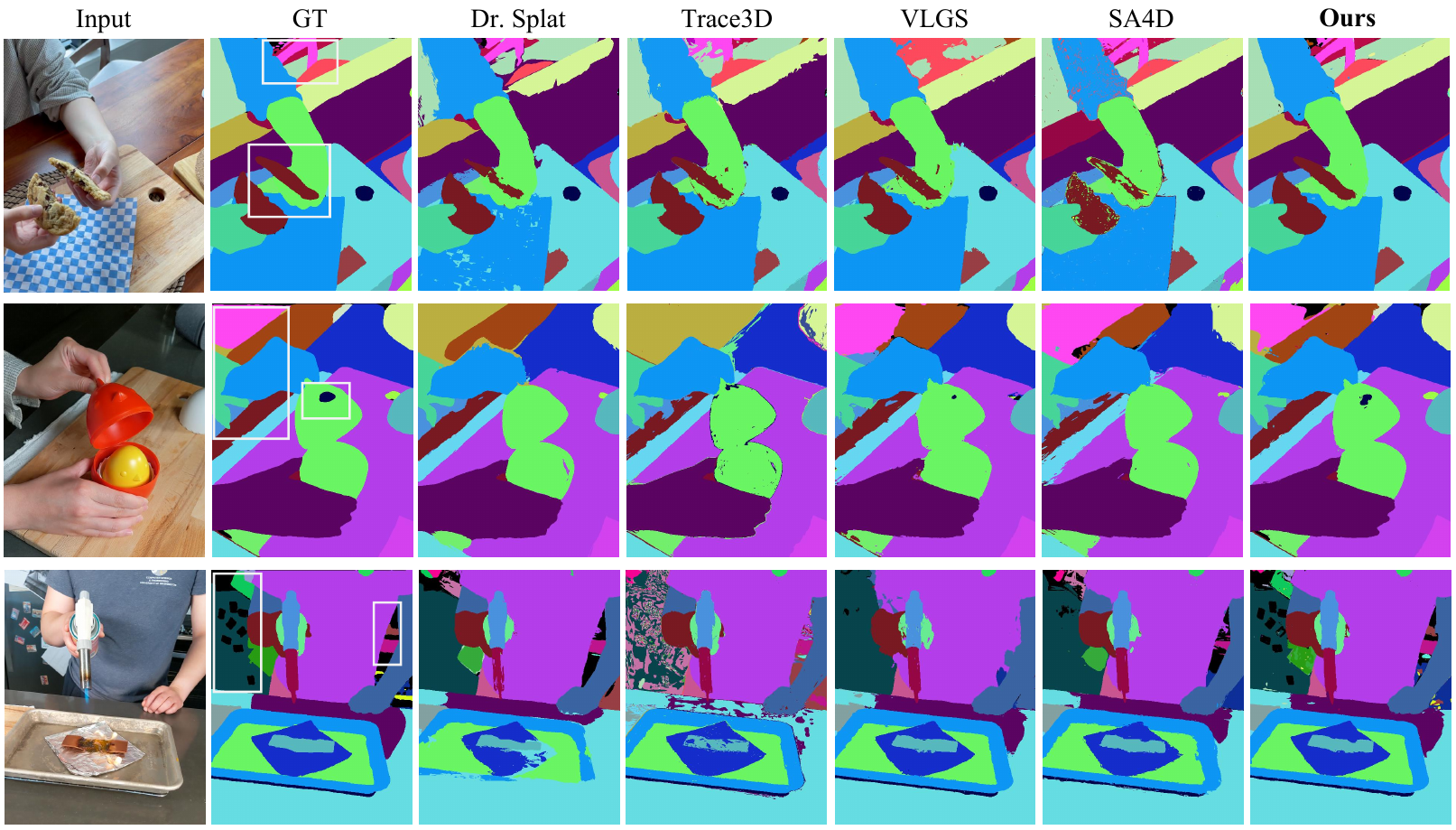}
  \vspace{-7mm}
  \caption{
      \textbf{Qualitative comparison of our method with the state-of-the-art on novel-view panoptic segmentation using the HyperNeRF~\cite{park2021hypernerf} dataset.} For clarity, we crop and slightly zoom in on representative regions around the manipulated objects. Our approach produces noticeably sharper and more coherent segmentations, even under occlusion and appearance variations.
  }
  \label{fig:hypernerf}
  \vspace{-3mm}
\end{figure*}

\noindent\textbf{Implementation Details.}
The Field-Aware Splatting module is implemented in CUDA, while other components are in PyTorch \cite{paszke2019pytorch}. All experiments are run on a single NVIDIA A40 GPU. Each scene is trained for 10,000 iterations for reconstruction and 3,000 iterations for instance segmentation using Adam \cite{kingma2014adam}.
Learning rates are 0.01 for occupancy and instance identity calibration, while other parameters use the default values from Deformable Gaussian Splatting \cite{wu20244d}. Instance-Guided Resampling uses a sampling rate of 1\% of all Gaussians for HyperNeRF and 5\% for Neu3D, and the instance loss weight ($\lambda_{\text{inst}}$) is set to 0.01 and 0.005, respectively. For mask rendering, we use argmax to obtain predicted instances without applying confidence thresholds.

\begin{table*}[t]
    \setlength{\tabcolsep}{3pt} 
    \centering
    \small
    \vspace{-1mm}
    \caption{
        \textbf{Quantitative comparison of our method with the state-of-the-art on novel-view panoptic segmentation using the Neu3D~\cite{li2022neural} dataset.}
        We report mAcc-pix, mAcc-inst, and mIoU metrics.
        The \colorbox{red!25}{best}, \colorbox{orange!25}{second best}, and \colorbox{yellow!25}{third best} results are highlighted.
    }
    \vspace{-2mm}
    \label{Table: neu3d}
    \resizebox{0.9\textwidth}{!}{%
    \small
    \begin{tabular}{l ccc c ccc c ccc}
        \toprule
        & \multicolumn{3}{c}{coffee martini}
        & & \multicolumn{3}{c}{cook spinach}
        & & \multicolumn{3}{c}{cut roasted beef} \\
        \cmidrule{2-4} \cmidrule{6-8} \cmidrule{10-12}
        Method
        & mAcc-pix & mAcc-inst & mIoU
        & & mAcc-pix & mAcc-inst & mIoU
        & & mAcc-pix & mAcc-inst & mIoU \\
        \midrule
        Dr. Splat~\cite{jun2025dr} & \colorbox{yellow!25}{88.37} & 73.84 & \colorbox{yellow!25}{70.74} && 78.46 & \colorbox{yellow!25}{76.07} & \colorbox{yellow!25}{69.55} && 53.55 & 30.03 & 25.96 \\
        Trace3D~\cite{shen2025trace3d} & 82.05 & \colorbox{yellow!25}{85.97} & 64.31 && 83.84 & 74.10 & 60.30 && 62.59 & 58.81 & 39.63 \\
        VLGS~\cite{peng20243d} & \colorbox{orange!25}{94.80} & \colorbox{orange!25}{94.08} & \colorbox{orange!25}{86.03} && \colorbox{orange!25}{95.60} & \colorbox{orange!25}{92.36} & \colorbox{orange!25}{85.74} && \colorbox{orange!25}{93.56} & \colorbox{orange!25}{83.36} & \colorbox{orange!25}{78.09} \\
        SA4D~\cite{ji2024segment} & 81.39 & 76.34 & 64.12 && \colorbox{yellow!25}{84.70} & 74.19 & 56.16 && \colorbox{yellow!25}{74.26} & \colorbox{yellow!25}{61.77} & \colorbox{yellow!25}{54.61} \\
        \textbf{Ours} & \colorbox{red!25}{96.07} & \colorbox{red!25}{95.04} & \colorbox{red!25}{91.50} && \colorbox{red!25}{96.63} & \colorbox{red!25}{93.82} & \colorbox{red!25}{87.61} && \colorbox{red!25}{95.12} & \colorbox{red!25}{85.78} & \colorbox{red!25}{80.24} \\
        \midrule
        & \multicolumn{3}{c}{flame salmon}
        & & \multicolumn{3}{c}{flame steak}
        & & \multicolumn{3}{c}{sear steak} \\
        \cmidrule{2-4} \cmidrule{6-8} \cmidrule{10-12}
        Method
        & mAcc-pix & mAcc-inst & mIoU
        & & mAcc-pix & mAcc-inst & mIoU
        & & mAcc-pix & mAcc-inst & mIoU \\
        \midrule
        Dr. Splat~\cite{jun2025dr} & \colorbox{yellow!25}{81.22} & \colorbox{yellow!25}{78.25} & \colorbox{orange!25}{74.32} && 69.13 & 62.97 & 57.34 && 73.72 & 74.88 & 66.79 \\
        Trace3D~\cite{shen2025trace3d} & \colorbox{orange!25}{86.25} & 76.20 & 67.97 && 73.62 & \colorbox{yellow!25}{89.63} & \colorbox{yellow!25}{77.04} && \colorbox{yellow!25}{86.99} & 90.30 & 78.14 \\
        VLGS~\cite{peng20243d} & 79.66 & \colorbox{orange!25}{82.83} & \colorbox{yellow!25}{70.83} && \colorbox{orange!25}{87.71} & \colorbox{orange!25}{95.28} & \colorbox{orange!25}{89.68} && \colorbox{orange!25}{90.20} & \colorbox{orange!25}{96.22} & \colorbox{orange!25}{84.60} \\
        SA4D~\cite{ji2024segment} & 78.89 & 75.26 & 60.79 && \colorbox{yellow!25}{81.54} & 80.36 & 66.29 && 86.93 & \colorbox{yellow!25}{94.57} & \colorbox{yellow!25}{80.12} \\
        \textbf{Ours} & \colorbox{red!25}{91.31} & \colorbox{red!25}{92.16} & \colorbox{red!25}{87.69} && \colorbox{red!25}{95.31} & \colorbox{red!25}{95.74} & \colorbox{red!25}{91.97} && \colorbox{red!25}{95.36} & \colorbox{red!25}{96.61} & \colorbox{red!25}{90.83} \\
        \bottomrule
    \end{tabular}
    }
\end{table*}

\begin{figure*}[t]
  \centering
  \includegraphics[width=1\textwidth]{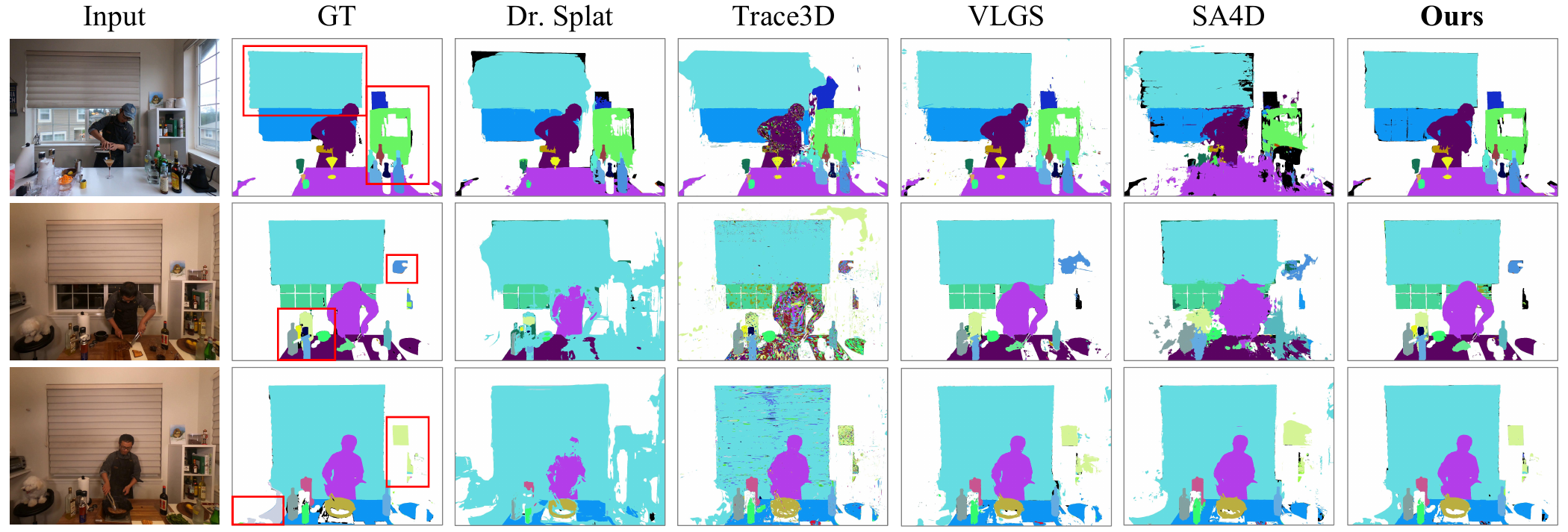}
  \vspace{-7mm}
  \caption{
      \textbf{Qualitative comparison of our method with the state-of-the-art on novel-view panoptic segmentation using the Neu3D~\cite{li2022neural} dataset.} As described in Sec.~\ref{sec:exp_setup}, to avoid the inherent inconsistencies in the ground truth, we consider only instances that are visible across all camera views. Our method produces smoother boundaries, cleaner backgrounds, and more consistent object identities.
  }
  \label{fig:neu3d}
  \vspace{-3mm}
\end{figure*}

\subsection{Comparison with the State-of-the-Art}
\label{sec:exp_compare_sota}
\noindent\textbf{Novel-View Panoptic Segmentation.}
Tables~\ref{Table: hypernerf} and ~\ref{Table: neu3d} compare our method with prior works~\cite{jun2025dr,peng20243d,shen2025trace3d,ji2024segment}. Across all scenes of both datasets, our method consistently outperforms existing approaches by a large margin. On the HyperNeRF~\cite{park2021hypernerf} dataset, our method achieves an average of 96.40 mAcc-pix, 85.69 mAcc-inst, and 79.47 mIoU, surpassing the second-best method (VLGS) by +2.09, +11.78, and +11.42, respectively. Notably, on the ``torchocolate'' scene, our method improves mIoU by +15.11 over SA4D and +20.89 over VLGS, demonstrating superior instance consistency and boundary accuracy. Similarly, on the Neu3D~\cite{li2022neural} dataset, our method achieves 94.97 mAcc-pix, 93.19 mAcc-inst, and 88.31 mIoU, outperforming VLGS by +4.72, +2.50, and +5.82, respectively. These consistent improvements across diverse dynamic scenes of both monocular and multi-view videos demonstrate the robustness and generalization ability of our proposed method.

\begin{figure*}[t]
  \centering
  \includegraphics[width=1\textwidth]{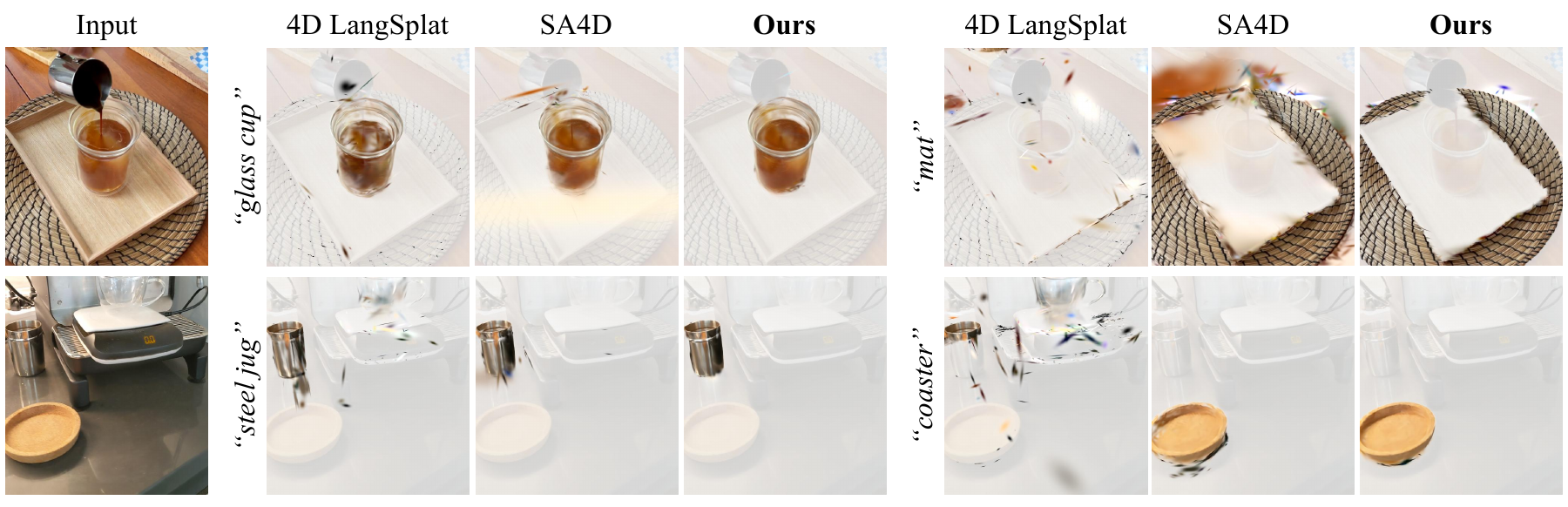}
  \vspace{-7mm}
  \caption{
   \textbf{Qualitative comparison of our method with the state-of-the-art on open-vocabulary 4D querying using the HyperNeRF~\cite{park2021hypernerf} dataset.} For clarity, we crop and zoom in on the central regions. Our method produces clearer boundaries and more accurate instance separation, even under transparent and reflective materials such as the glass cup and steel jug.
   }
  \label{fig:open}
  \vspace{-3mm}
\end{figure*}

\begin{table}[t]
\centering
\small
\caption{\textbf{Ablation study.} We evaluate our method under different configurations on the ``split-cookie'' scene from HyperNeRF~\cite{park2021hypernerf}.}
\label{tab:ablation1}
\vspace{-2mm}
\resizebox{\linewidth}{!}{%
\begin{tabular}{lcccc}
\toprule
Method & mAcc-pix & mAcc-inst & mIoU & PSNR \\
\midrule
(i) Const. Occ. & 96.26 & 85.57 & 80.80 & 31.76 \\
(ii) Opa. Occ. & 96.60 & 87.20 & 82.34 & 32.16 \\
(iii) w/o Calib. & 95.99 & 82.65 & 78.16 & 26.73 \\
(iv) w/o Resamp. & 96.78 & 87.98 & 82.82 & 32.34 \\
\textbf{(v) Full} & \textbf{97.93} & \textbf{90.40} & \textbf{86.03} & \textbf{32.42} \\
\bottomrule
\end{tabular}
}
\vspace{-3mm}
\end{table}

\begin{figure}[t]
  \centering
  \includegraphics[width=1\linewidth]{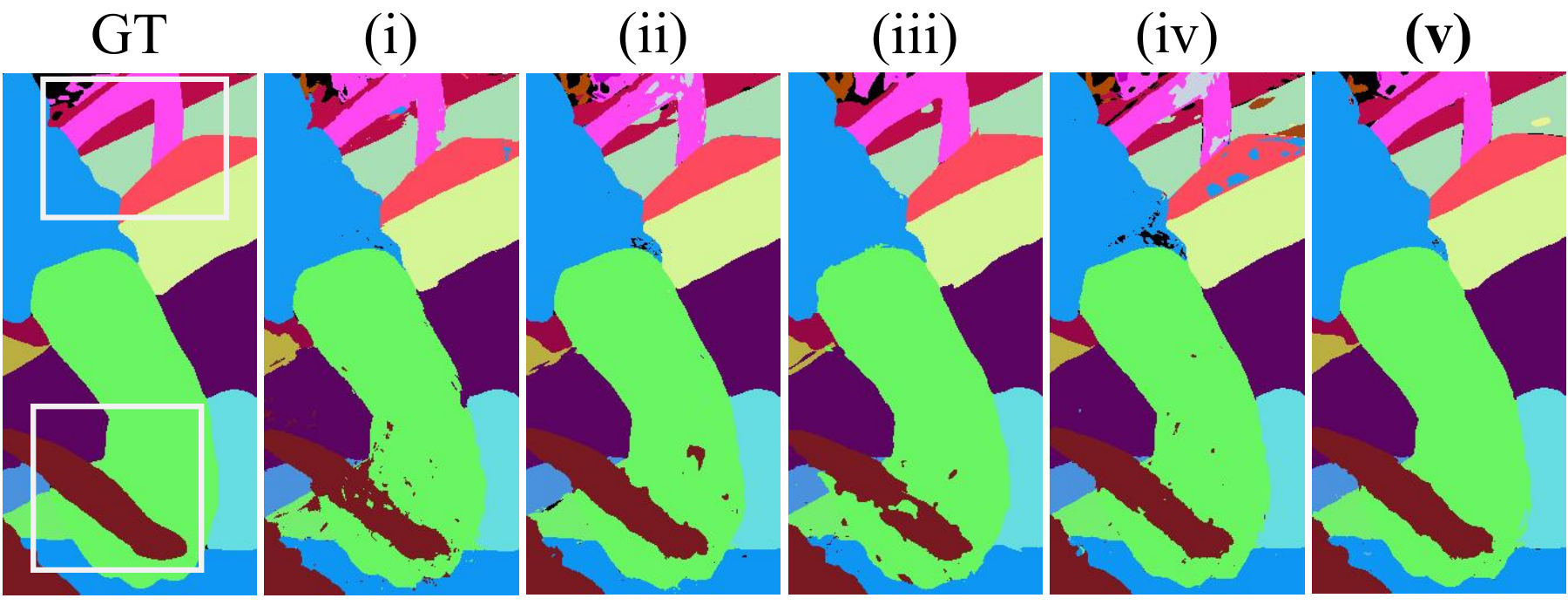}
  \vspace{-6mm}
  \caption{
    \textbf{Ablation study.} We present the corresponding qualitative results for each configuration shown in Table~\ref{tab:ablation1}.
  }
  \label{fig:ablation}
  \vspace{-5mm}
\end{figure}

Figures~\ref{fig:hypernerf} and~\ref{fig:neu3d} further provide qualitative comparisons, which highlight the strength of the Consistent Instance Field in preserving spatial precision and semantic coherence in dynamic scenes.
As illustrated in Figure~\ref{fig:hypernerf}, on the HyperNeRF dataset, our method produces finer segmentation with clearer instance separation, especially around hand-object interactions and partial occlusions, where prior methods yield fragmented or flickering masks. Similarly, as shown in Figure~\ref{fig:neu3d}, on the Neu3D dataset, our method maintains precise object boundaries and consistent identities, even in cluttered environments.
More qualitative video results are provided in the supplementary material.

\noindent\textbf{Open-Vocabulary 4D Querying.}
Figure~\ref{fig:open} presents qualitative comparisons on open-vocabulary 4D querying against the recent state-of-the-art method for dynamic scene understanding, 4D LangSplat~\cite{li20254d}.
For comprehensive evaluation, we also extend SA4D~\cite{ji2024segment} to this task by using Grounded DINO~\cite{liu2024grounding} for text-mask correspondence.
Our approach achieves more accurate instance localization and sharper boundaries, even under challenging visual conditions involving transparent or reflective objects (\eg, the glass cup and steel jug). In contrast, both baselines suffer from boundary leakage or semantic confusion.
Additional video and quantitative results are provided in the supplementary material.

\subsection{Ablation Studies}
\label{sec:exp_ablation}
We conduct controlled ablations to assess the impact of each component in our proposed method in Table~\ref{tab:ablation1}. Each variant modifies a single module while keeping all other settings identical to the full method, ensuring fair and isolated comparisons. We consider five configurations: \textbf{(i) Constant Occupancy}, where occupancy ($\pi_i$) is fixed to 0.02, removing spatial adaptivity; \textbf{(ii) Opacity as Occupancy}, using RGB opacity as a surrogate for occupancy, which blurs the distinction between existence and visibility; \textbf{(iii) w/o Identity Calibration}, removing the calibration term ($m_i^k$), allowing 2D visibility bias to propagate into the 4D instance distribution; \textbf{(iv) w/o Instance-Guided Resampling}, disabling resampling so Gaussian capacity may be redundant in background but insufficient in active regions; and \noindent\textbf{(v) Full}, our complete approach.
The results show that removing calibration (iii) or resampling (iv) significantly degrades performance, producing inconsistent and noisy segmentation. Using opacity as occupancy (ii) retains coarse geometry but fails to preserve instance consistency, while constant occupancy (i) also reduces performance due to lack of spatial adaptivity. In contrast, the full method (v), which jointly learns adaptive occupancy, calibrated identities, and resampled Gaussians, achieves the best balance of geometric fidelity and semantic consistency. Figure~\ref{fig:ablation} qualitatively illustrates that only the full method maintains coherent segmentation through complex hand-object interactions.

\section{Conclusion}
\label{sec:conclusion}
We presented the Consistent Instance Field, a unified probabilistic framework for dynamic scene understanding that jointly models geometry, motion, and semantics in a continuous 4D representation. CIF disentangles visibility from identity and leverages instance-field signals to adaptively organize Gaussians around semantically meaningful entities, improving both spatial accuracy and temporal coherence. We demonstrate that CIF significantly outperforms existing methods on both novel-view panoptic segmentation and open-vocabulary 4D querying tasks. For future work, we aim to introduce a standardized 4D evaluation benchmark for open-vocabulary 4D querying, enabling more consistent assessment and driving further advances in 4D instance-level modeling.


{
    \small
    \bibliographystyle{ieeenat_fullname}
    \bibliography{main}
}

\appendix
\beginsupplement
\maketitlesupplementary

\begin{figure*}[!b]
  \centering
  \includegraphics[width=1\textwidth]{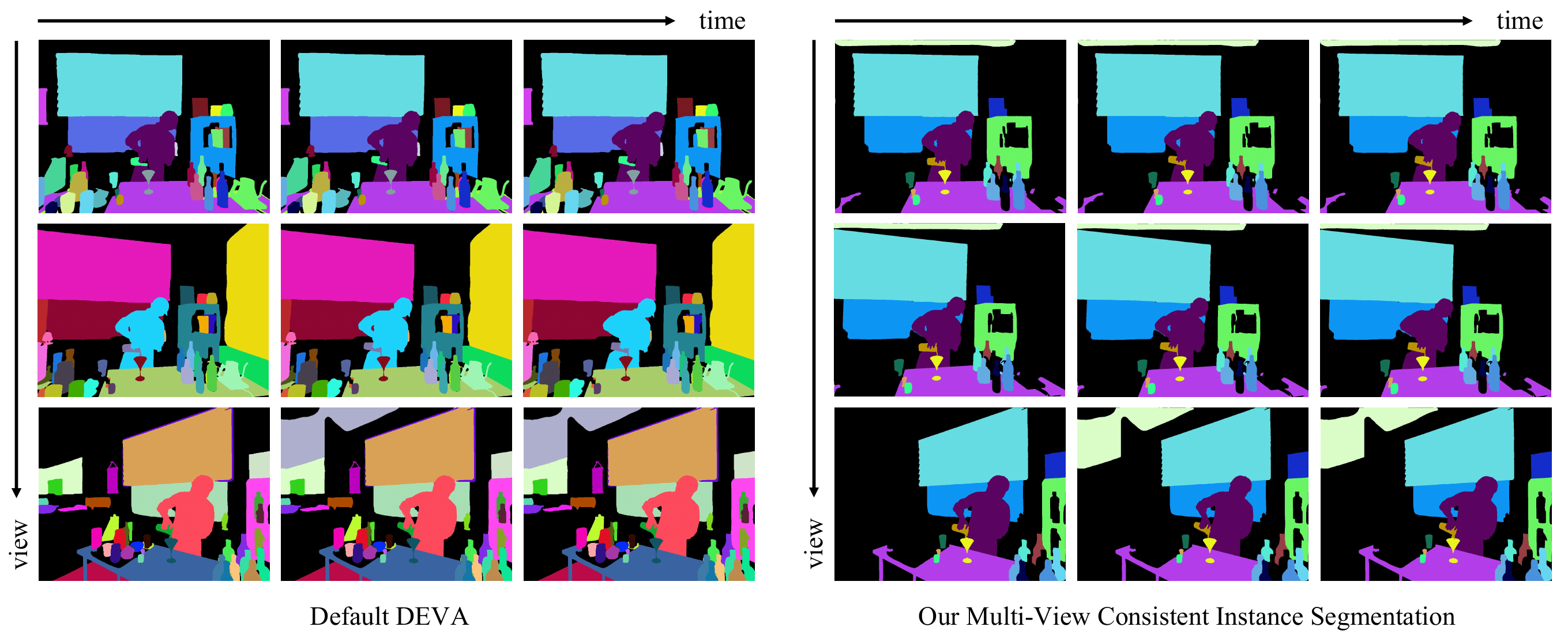}
  \vspace{-6mm}
  \caption{
  \textbf{Results of multi-view video segmentation on Neu3D \cite{li2022neural} dataset.}
  (\textbf{Left}) DEVA~\cite{cheng2023tracking} produces instance masks independently per camera view, leading to inconsistent instance identities across synchronized camera streams, (\textbf{Right}) our multi-view video segmentation results by merging multi-videos into a single pseudo-monocular sequence to produce multi-view consistent instance masks.
 }
  \label{fig:cross-view}
\end{figure*}

\begin{table*}[!t]
    \setlength{\tabcolsep}{6pt}
    \centering
    \small
    \caption{
        \textbf{Quantitative comparison of our method with the state-of-the-art on open-vocabulary 4D querying using the HyperNeRF \cite{park2021hypernerf} dataset.}
        We report mAcc and mIoU metrics.
        The \colorbox{red!25}{best}, \colorbox{orange!25}{second best}, and \colorbox{yellow!25}{third best} results are highlighted.
        * indicates failure of localizing the objects based on the text queries, as also demonstrated in Figure 5 of the main paper.
    }
    \vspace{-1mm}
    \label{tab:supp:ov}
    \resizebox{0.8\textwidth}{!}{
    \begin{tabular}{l c c c c c c c c}
        \toprule
        & \multicolumn{4}{c}{americano} 
        & \multicolumn{4}{c}{espresso} \\
        \cmidrule(lr){2-5} \cmidrule(lr){6-9}
        & \multicolumn{2}{c}{\textit{``glass cup''}} & \multicolumn{2}{c}{\textit{``mat''}}
        & \multicolumn{2}{c}{\textit{``steel jug''}} & \multicolumn{2}{c}{\textit{``coaster''}} \\
        \cmidrule(lr){2-3} \cmidrule(lr){4-5} 
        \cmidrule(lr){6-7} \cmidrule(lr){8-9}
        Method & mAcc & mIoU & mAcc & mIoU
        & mAcc & mIoU & mAcc & mIoU \\
        \midrule

        4D LangSplat~\cite{li20254d}
        & \colorbox{orange!25}{98.02} & \colorbox{orange!25}{78.17}
        & \colorbox{orange!25}{72.90} & \colorbox{yellow!25}{7.55*}
        & \colorbox{yellow!25}{96.33} & \colorbox{yellow!25}{35.61}
        & \colorbox{yellow!25}{86.69} & \colorbox{yellow!25}{0.43*} \\

        SA4D~\cite{ji2024segment}
        & \colorbox{yellow!25}{88.91} & \colorbox{yellow!25}{40.05}
        & \colorbox{yellow!25}{71.65} & \colorbox{orange!25}{39.67}
        & \colorbox{orange!25}{99.25} & \colorbox{orange!25}{70.49}
        & \colorbox{orange!25}{99.56} & \colorbox{orange!25}{81.12} \\

        \textbf{Ours}
        & \colorbox{red!25}{99.02} & \colorbox{red!25}{88.52}
        & \colorbox{red!25}{94.68} & \colorbox{red!25}{77.66}
        & \colorbox{red!25}{99.72} & \colorbox{red!25}{87.93}
        & \colorbox{red!25}{99.73} & \colorbox{red!25}{85.47} \\

        \bottomrule
    \end{tabular}
    }
\end{table*}

\noindent\textbf{Overview.}
In the supplementary material, we first present our proposed multi-view instance segmentation to get cross-view consistent instance masks on the Neu3D dataset (Sec. \ref{sec:supp:cross-view}). We then present additional qualitative video results (Sec. \ref{sec:supp:qual}) and quantitative results (Sec. \ref{sec:supp:quan}). Finally, we discuss the limitation of our method (Sec. \ref{sec:supp:limit}) and the societal impact (Sec. \ref{sec:supp:impact}).

\section{Implementation Details}\label{sec:supp:detail}
\subsection{Multi-View Consistent Instance Segmentation}\label{sec:supp:cross-view}
\noindent\textbf{Pre-processing (merging multi-view videos).} As discussed in Sec. 4.1 of the main paper, the multi-view benchmark Neu3D \cite{li2022neural} provides synchronized videos but does not include ground-truth instance annotations. Therefore, we follow prior works \cite{ji2024segment, li20254d} and use DEVA \cite{cheng2023tracking}, a video object tracking model, to generate input masks. However, because DEVA processes each video independently, the resulting masks are inconsistent across views. To address this limitation, we reinterpret the spatial multi-view dimension as an inter-temporal one and merge all views into a single pseudo-monocular sequence. More precisely, we concatenate each video with the reversed video of its adjacent view. Formally, given \(N\) spatially adjacent camera views and \(T\) frames per video corresponding to each view, we first reorder frames to maximize temporal continuity:
\begin{equation}
\mathcal{S}_{n}^{\text{ordered}} =
\begin{cases}
(I_1, I_2, \dots, I_T), & \text{if } n \text{ is odd},\\[2mm]
(I_T, I_{T-1}, \dots, I_1), & \text{if } n \text{ is even},
\end{cases}
\label{eq:zigzag}
\end{equation}
where \(I_t\) denotes the \(t\)-th frame from that $n$-th view. The input to DEVA is then constructed by concatenating all reordered view sequences:
\begin{equation}
\mathcal{S}_{1:N}^{\text{ordered}} = \big[\,\mathcal{S}_{1}^{\text{ordered}},\,\mathcal{S}_{2}^{\text{ordered}},\,\dots,\,\mathcal{S}_{N}^{\text{ordered}}\,\big].
\end{equation}
This merged pseudo-monocular sequence ensures that adjacent frames vary smoothly across both time and viewpoint.

\noindent\textbf{Video instance segmentation.}
Given a concatenated video $\mathcal{S}_{1:N}^{\text{ordered}}$, DEVA then produces per-frame instance masks with temporally propagated instance IDs.

\noindent\textbf{Post-processing (visibility filtering).} 
Due to occlusions and limited camera overlap, some instances may disappear entirely in certain views, leading to conflicting identities across cameras. To prevent incomplete or inconsistent masks, we retain only instances that remain visible in all views. Specifically, an instance $k$ is considered valid if its mask has non-empty support in every view of the concatenated videos. Instances that do not meet this criterion are discarded.

Figure \ref{fig:cross-view} compares the default DEVA results, where each video is segmented independently, with our multi-view consistent instance segmentation. This strategy effectively converts the multi-view videos into a single coherent identity sequence, enabling cross-view consistent pseudo-labels for both supervision and evaluation.

\section{Experimental Results}\label{sec:supp:result}

\subsection{Additional Qualitative Results}\label{sec:supp:qual}
To further demonstrate that our Consistent Instance Field provides coherent instance understanding across both space and time, we include additional qualitative video results comparing our method with recent state-of-the-art approaches SA4D~\cite{ji2024segment}, Trace3D~\cite{shen2025trace3d}, Dr.Splat~\cite{jun2025dr}, and VLGS~\cite{peng20243d} on the standard HyperNeRF~\cite{park2021hypernerf} and Neu3D~\cite{li2022neural} benchmarks for both novel-view panoptic segmentation and open-vocabulary 4D querying tasks:
\begin{itemize}
    \item Results on the monocular benchmark HyperNeRF~\cite{park2021hypernerf} for novel-view panoptic segmentation:\\
    \texttt{panoptic\_segmentation\_hypernef.mp4}
    \item Results on the multi-view benchmark Neu3D~\cite{li2022neural} for novel-view panoptic segmentation:\\\texttt{panoptic\_segmentation\_neu3d.mp4};
    \item Results on the monocular benchmark HyperNeRF~\cite{park2021hypernerf} for open-vocabulary 4D querying:\\\texttt{open\_vocabulary\_4d\_querying\_hypernerf.mp4}
\end{itemize}

\subsection{Additional Quantitative Results}\label{sec:supp:quan}
To further assess the capability of the Consistent Instance Field in open-vocabulary 4D querying, we report additional quantitative results using standard metrics, mAcc and mIoU.
We follow the experimental setting described in the main paper and use Grounded DINO \cite{liu2024grounding} to obtain 2D masks for each text query as pseudo-ground-truth annotations.

As shown in Table~\ref{tab:supp:ov}, our method consistently achieves higher retrieval accuracy across various text prompts.
On average, our method achieves 98.29 mAcc and 84.90 mIoU, surpassing the second-best approach (SA4D) by 8.45 and 27.07, respectively.
In contrast, previous methods like 4D LangSplat \cite{li20254d} often struggle to localize the objects based on the text query, as demonstrated in Figure 5 of the main paper, thus yielding extremely low performance (\eg, 7.55 mIoU on the ``americano'' scene with the ``mat'' query and 0.43 on the ``espresso'' scene with the ``coaster'' query).
These results highlight the strength of our method in fine-grained and coherent instance modeling in complex dynamic scenes.

\section{Discussions}\label{sec:supp:discussion}

\subsection{Limitations}\label{sec:supp:limit}
While the proposed Consistent Instance Field provides a principled formulation for identity modeling in dynamic scenes, several limitations remain.
First, our formulation is instantiated via a deformable Gaussian representation \cite{yang2024deformable, wu20244d}, which inherits its representational constraints.
Scenes involving amorphous or continuously evolving materials (\eg, smoke or liquids) lack stable structure and may not be faithfully represented through persistent Gaussian primitives.
In such cases, identity assignments become less interpretable, as the Gaussians fail to maintain consistent spatial support or correspond to physically persistent entities.
Future advances in dynamic scene representations may help extend our method to broader scene types.
Second, although we construct pseudo-monocular sequences to synchronize multi-view pseudo-labels from video object tracking models \cite{cheng2023tracking}, residual cross-view inconsistencies or missing annotations under severe occlusion can still bias identity estimation. Exploring more rigorous multi-view pseudo-label harmonization strategies represents a promising direction for enhancing robustness in such scenarios.

\subsection{Societal Impact}\label{sec:supp:impact}
Our approach provides structured, instance-consistent scene representations that extend beyond visual reconstruction to support simulation, prediction, and interaction within dynamic environments.
These advances could improve the safety, efficiency, and interpretability of autonomous and interactive systems, provided that ethical and privacy standards are respected.


\end{document}